\pdfoutput=1

\documentclass[11pt]{article}

\usepackage{acl}

\usepackage{times}
\usepackage[]{mdframed}
\usepackage{latexsym}
\usepackage{amsmath}
\usepackage{physics}
\usepackage{bbding}
\usepackage{pifont}
\usepackage{wasysym}
\usepackage{amssymb}
\usepackage[T1]{fontenc}

\usepackage[utf8]{inputenc}

\usepackage{microtype}

\usepackage{inconsolata}

\usepackage{graphicx}
\usepackage{xcolor}
\usepackage{soul}
\usepackage{times}
\usepackage{latexsym}
\usepackage{multirow}
\usepackage{graphicx}
\usepackage{makecell}
\usepackage{booktabs}
\usepackage{colortbl}
\definecolor{aqua}{rgb}{0.0, 1.0, 1.0}
\definecolor{hot pink}{rgb}{1.0, 0.31, 0.61}

\usepackage{xcolor,pifont}
\newcommand*\colourcheck[1]{%
  \expandafter\newcommand\csname #1check\endcsname{\textcolor{#1}{\ding{52}}}%
}
\title{Dictionary Insertion Prompting for Multilingual Reasoning on\\Multilingual Large Language Models}

\author{
    Hongyuan Lu$^\heartsuit$\Thanks{\hspace{1mm}Equal Contribution.}, Zixuan Li
$^{\spadesuit*}$, Wai Lam$^\heartsuit$\\
    $\heartsuit$The Chinese University of Hong Kong\\
    $^\spadesuit$Cyber Science and Engineering, Southeast University\\
    \{hylu,wlam\}@se.cuhk.edu.hk\\
    zixuan.li@seu.edu.cn 
}

\begin{document}
\maketitle
\begin{abstract}
There are two shortages in the current Large Language Models (LLMs) era. The first is short of multilingual models, where most LLMs are English-centric and performance is limited on multilingual reasoning. The second is the place of external knowledge to be used, where most retrieved knowledge is prepended to the user queries (maybe sub-optimal). This paper presents a novel and simple yet effective method called \textbf{D}ictionary \textbf{I}nsertion \textbf{P}rompting (\textbf{DIP}). When providing a non-English prompt, DIP looks up a word dictionary and inserts words' English counterparts into the middle of the prompt for LLMs. It then enables better translation into English and better English model thinking steps which leads to obviously better results. We experiment with 10 to 200 languages from FLORES-200.\footnote{The number of languages varies on the datasets, and we experiment with 200 languages on GSM8K as in Appendix} Since there are no adequate datasets, we use the NLLB translator to create synthetic multilingual benchmarks from the existing 4 English reasoning benchmarks such as GSM8K and AQuA. The synthetic benchmarks are translated back into English for quality assurance with manual annotation. Interestingly, the place for injecting the dictionary plays an important factor in the performance gains, and we found that interleaving the dictionary with the original words gives a better performance compared to prepending/appending the dictionary, under the same dictionary constructed.
\end{abstract}

 \section{Introduction}
 In the quick development with large language models  (LLMs), there have been quite many popular research areas such as chain-of-thought reasoning \citep{wang-etal-2023-towards,10.5555/3600270.3602070,yang-etal-2024-unveiling}, machine translation \citep{2023arXiv230506575L,zhu-etal-2024-multilingual,zhu-etal-2024-clean}, code generation \citep{li-etal-2023-codeie,zhang-etal-2023-self,hou-etal-2025-lne}, dialogue generation \citep{li-etal-2022-grounded, lu-etal-2022-partner, lu-lam-2023-pcc,yang-etal-2025-stephanie,2026arXiv260105657Y}, and even spatial understanding \citep{hu2024chainofsymbol}. Among these, an important research area is multilingual large language models (MLLMs), which consider not only the tasks of machine translation but also reasoning tasks represented in different languages \citep{huang-etal-2023-languages}. This scales the horizon of English-centric LLMs such as popular ChatGPT and enables them to be used by people who mainly speak low-resourced languages. Yet, current methods are usually training-based \citep{2024arXiv240705975L,2024arXiv240913949L}, which usually requires many GPU/TPU computational resources to update model weights from LLMs. 
 \par
 \begin{figure*}[t!]
\begin{center}
\vspace{0mm}
\centerline{
\includegraphics[width=16cm]{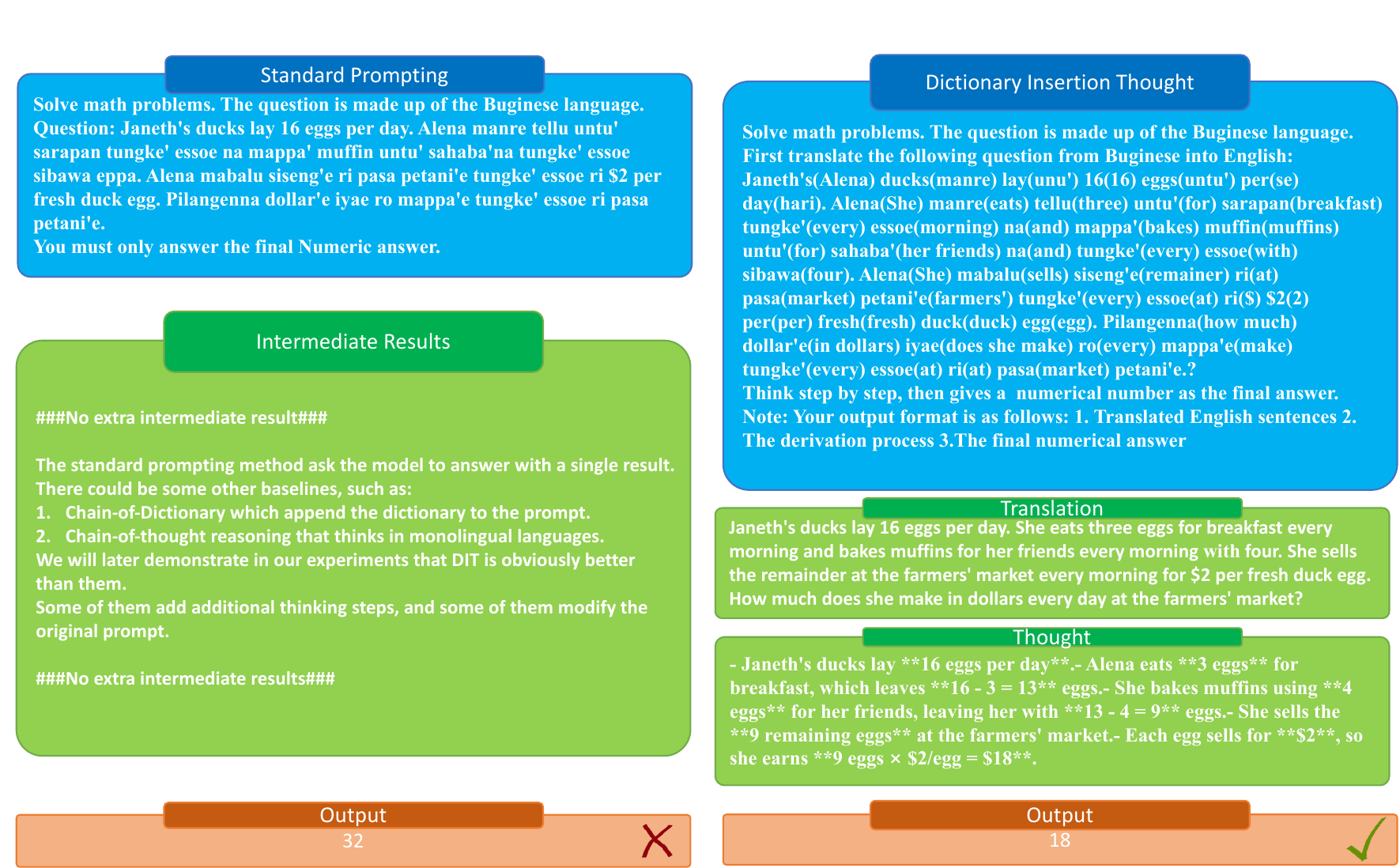}}
\caption{An illustrated comparison of the GSM8K dataset made up in Buginese. Compared to the standard prompting baseline, DIP inserts additional dictionary knowledge written in knowledge in an interleaving manner. This leads to a better intermediate English translation and succeeding English thought. Finally, DIP produces promising results, surpassing several strong baselines.}
\label{dit_f}
\end{center}
\vspace{-5mm}
\end{figure*}
 In contrast, this paper investigates how to incorporate a dictionary as an auxiliary knowledge into prompting. In comparison, this is lightweight and flexible, where the dictionary can be customized and replaced in a plug-and-play manner without model training. While the dictionary-based method has been studied on the task of traditional machine translation \citep{arthur-etal-2016-incorporating}, how to incorporate them into reasoning tasks on MLLMs has been under-studied. This is yet important and needs to be empirically justified whether, and how dictionary-based methods can be better used to improve multilingual reasoning tasks, which obviously enhances LLMs' usefulness in our daily life.
 \par 
 To this end, we propose a novel method called \textbf{D}ictionary \textbf{I}nsertion \textbf{P}rompting (\textbf{DIP}). We present the overall algorithm as in Figure \ref{dit_f}. By providing a customized dictionary that maps words represented in low-resourced languages into English, DIP inserts the English representation into the original input. This then helps LLMs in pivoting the original input into a complete English representation. This improves the succeeding chain of thought reasoning, which results in a better final output. 
 \par
 By looking deeply into the experimental analysis, we found that the place of insertion is an important factor in improving the performance of DIP. While the prior dictionary-based method usually presents the dictionary in front of the prompt, we found it sub-optimal, and placing the dictionary in an interleaving manner is better. We postulate that such a design puts words with their English counterparts closer, and makes it easy to be understood by LLMs. Experiments also show that better translation quality and thought quality are the key factors to the usefulness of DIP. 
 \par
 We benchmark DIP on about 200 languages from FLORES-200 \citep{nllb2022}. Since there are no adequate datasets for covering those languages, we use a high-quality SOTA translator NLLB 3.3B\footnote{https://huggingface.co/spaces/Narrativaai/NLLB-Translator} to translate existing arithmetic reasoning benchmarks such as GSM8K \citep{2021arXiv211014168C} and commonsense reasoning benchmarks such as Date \citep{2022arXiv220604615S}.
 \par
 In general, we make three key contributions.
\begin{itemize}
\setlength\itemsep{0em}
    \item We propose a simple, novel, yet computationally lightweight approach called DIP for better reasoning on multilingual tasks on LLMs.
    \item Extensive strong results across several benchmarks on ChatGPT and Llama LLMs verify the effectiveness of DIP.  
    \item We investigated further why DIP is useful and found that better intermediate translation and thinking steps are the key.
\end{itemize}
\section{Dictionary Insertion Prompting}
Large language models show their promising translation performance when sufficiently pre-trained \citep{2023arXiv230304048W,lu-etal-2023-trip, tang-etal-2024-metrics, lu-etal-2024-revamping, lu-etal-2024-chain, lu-etal-2025-slow, 2026arXiv260402176L}. Yet, such translation ability usually diminishes in low-resourced languages and it is still an under-studied topic for reasoning in those low-resourced languages such as the ones from FLORES-200 \citep{nllb2022}.
\par
This paper proposes a novel, yet simple and effective framework called \textbf{DIP} (\textbf{D}ictionary \textbf{I}nsertion \textbf{P}rompting) to address these difficulties by integrating the dictionary knowledge into the reasoning process. DIP first looks up a customized dictionary from the original non-English prompt to place its English counterpart accordingly. This is followed by pivoting into English, which succeedingly invokes a better English thinking process. The final result is then obviously improved by DIP.
\par
Therefore, DIP is as illustrated in Figure \ref{dit_f}:
\begin{mdframed}
(1) Solve the question. The question is made up of the <language> language.
\\
(2) First translate the following question from <language> into English: 
\\
(3) <question>\\
(4) Note: Your output format is as follows: \\1. Translated English sentences \\2. The derivation process \\3.The final numerical answer
\end{mdframed}
, where <language> denotes the language of the non-English question prompts\footnote{We conduct our experiments on non-English languages with English-centric LLMs, and we leave other settings to future work.} that are written in, and <question> denotes the actual question that is written in those non-English languages. 
\par 
The reason for choosing English as the pivoting language for DIP on LLMs is due to the fact that English has been primarily used as the pivoting language in traditional machine translation \citep{utiyama-isahara-2007-comparison,wu-wang-2007-pivot}. While other languages can be possibly useful, English is the most high-resourced language on English-centric LLMs, it is intuitively the most helpful, so other attempts are left to future works. 
\paragraph{Dictionary Construction} To construct the bilingual dictionary mapping between English and the original prompt, we prompt ChatGPT:
\begin{mdframed}
(1) Please provide the translation of the given English sentence into <language>, along with a word-for-word dictionary for each word.\\
(2) The output format must be strictly followed: \\1. Start with `English:' followed by the English sentence. \\2. On the next line, start with `<lang>:' followed by the <language> translation. \\3. On the next line, start with `dictionary:' followed by each word in the <language> sentence, annotated with its English meaning in parentheses, separated by spaces.
\\
(3) Now generate translations for the following sentence: 
\\
English: <target>
\\
<language>: <source>
\\ 
dictionary:
\end{mdframed}
, where <language> presents the language that the original synthetic questions are written in, <target> represents the original English sentence which is used to obtain the <source> sentences that are written in <language>.
\par
By prompting LLMs, we obtain a dictionary mapping that could be customized by replacing the arguments with any bilingual corpora.
\par 
Note that the pivoting into English and English-based thoughts is indeed optional in DIP and can be pruned for a better trade-off between performance and computational cost, as longer generation usually requires more computational costs.
\section{Experimental Setup}
\subsection{Baselines}
We conduct experiments with ChatGPT (GPT-4o-mini), Llama-3.2-1b-instruct \citep{2024arXiv240721783D}, and Mixtral-8x7b \citep{2024arXiv240104088J}. At the time of writing, all of them are popular and widely used English-centric LLMs which are strong in their multilingual and reasoning capacities. Based on these popular LLMs, we compare DIP to strong baseline methods:
\begin{table*}[thb!]
\scriptsize
\centering
\setlength\tabcolsep{3pt}
\setlength\extrarowheight{4pt}
\begin{tabular}{lcccccccccc|c}
\hline
\noalign{\vskip 1mm}  
\textbf{Model} & \textbf{kaz\_Cyrl} & \textbf{nso\_Latn} & \textbf{srp\_Cyrl} & \textbf{xho\_Latn}& \textbf{ibo\_Latn}& \textbf{tum\_Latn}& \textbf{asm\_Beng}& \textbf{bug\_Latn}& \textbf{ckb\_Arab}&     \textbf{azb\_Arab} & \textbf{Average}\\
\noalign{\vskip 1mm}  
\hline
\hline
\noalign{\vskip 1mm}  
Standard Prompting&17.89&9.10&15.62&11.52&11.30&5.91&14.25&5.84&9.55&9.40&11.04\\
Non-insertion Prompting&14.94&6.29&13.87&10.08&10.24&6.37&12.89&7.05&10.24&8.19&10.02\\
English Pivoting&23.65&14.03&19.94&19.79&19.26&10.31&21.91&9.55&17.82&17.44&17.37\\
English Pivot Thought&61.11&36.77&60.35&57.16&49.13&22.67&60.96&20.85&44.28&35.48&44.88\\
\noalign{\vskip 1mm}  
\hline
\hline
\noalign{\vskip 1mm}  
DIP w/o EP w/o CT &20.39&12.05&22.06&15.47&13.57&12.43&19.03&13.95&16.38&18.20&16.35\\
DIP w/ EP w/o CT &23.43&16.30&24.94&21.23&18.50&14.86&23.58&19.18&22.14&22.37&20.65\\
DIP &\textbf{67.93}&\textbf{46.17}&\textbf{80.36}&\textbf{67.10}&\textbf{53.30}&\textbf{43.29}&\textbf{68.61}&\textbf{60.50}&\textbf{63.68}&\textbf{68.31}&\textbf{61.92}\\
\noalign{\vskip 1mm}  
\hline
\end{tabular}
\caption{\label{t1}
Results for GPT-4o on GSM8K on 10 randomly selected low-resourced languages from FLORES-200. EP denotes the English pivoting translation process, and CT denotes chain-of-thought reasoning steps.
}
\end{table*}
\begin{table*}[thb!]
\scriptsize
\centering
\setlength\tabcolsep{3pt}
\setlength\extrarowheight{4pt}
\begin{tabular}{lcccccccccc|c}
\hline
\noalign{\vskip 1mm}  
\textbf{Model} & \textbf{kaz\_Cyrl} & \textbf{nso\_Latn} & \textbf{srp\_Cyrl} & \textbf{xho\_Latn}& \textbf{ibo\_Latn}& \textbf{tum\_Latn}& \textbf{asm\_Beng}& \textbf{bug\_Latn}& \textbf{ckb\_Arab}&     \textbf{azb\_Arab} & \textbf{Average}\\
\noalign{\vskip 1mm}  
\hline
\hline
\noalign{\vskip 1mm}  
Standard Prompting&53.33&42.00&46.33&48.33&36.67&28.33&53.00&25.33&37.67&40.67&41.17\\
Non-insertion Prompting&52.67&34.33&45.00&46.00&38.33&29.67&53.00&28.67&45.67&41.33&41.47\\
English Pivoting&54.67&51.33&51.67&63.00&53.67&40.33&63.67&35.00&57.33&51.67&52.23\\
English Pivot Thought&61.33&\textbf{61.33}&58.67&71.00&60.67&41.67&71.00&36.67&69.33&55.33&58.70\\
\noalign{\vskip 1mm}  
\hline
\hline
\noalign{\vskip 1mm}  
DIP w/o EP w/o CT &66.00&43.00&73.00&65.67&49.00&45.33&62.67&48.67&62.33&61.67&57.73\\
DIP w/ EP w/o CT &66.67&55.67&75.67&73.67&57.00&55.67&70.00&62.00&67.33&66.00&64.97\\
DIP &\textbf{78.33}&57.00&\textbf{89.67}&\textbf{76.67}&\textbf{65.33}&\textbf{65.67}&\textbf{77.67}&\textbf{71.00}&\textbf{78.67}&\textbf{75.00}&\textbf{73.50}\\
\noalign{\vskip 1mm}  
\hline
\end{tabular}
\caption{\label{t2}
Results for GPT-4o on SVAMP on 10 randomly selected low-resourced languages from FLORES-200. EP denotes the English pivoting translation process, and CT denotes chain-of-thought reasoning steps.
}
\end{table*}
\begin{table*}[thb!]
\scriptsize
\centering
\setlength\tabcolsep{3pt}
\setlength\extrarowheight{4pt}
\begin{tabular}{lcccccccccc|c}
\hline
\noalign{\vskip 1mm}  
\textbf{Model} & \textbf{kaz\_Cyrl} & \textbf{nso\_Latn} & \textbf{srp\_Cyrl} & \textbf{xho\_Latn}& \textbf{ibo\_Latn}& \textbf{tum\_Latn}& \textbf{asm\_Beng}& \textbf{bug\_Latn}& \textbf{ckb\_Arab}&     \textbf{azb\_Arab} & \textbf{Average}\\
\noalign{\vskip 1mm}  
\hline
\hline
\noalign{\vskip 1mm}  
Standard Prompting & 4.00 & 1.00 & 3.67 & 3.00 & 3.00 & 3.67 & 3.00 & 2.67 & 3.33 & 3.00 & 3.03 \\

Non-insertion Prompting & 2.67 & 2.00 & 3.00 & 3.00 & 2.67 & 1.33 & 1.67 & 3.00 & 1.00 & 1.00 & 1.87 \\

English Pivoting & 4.00 & 2.67 & 9.00 & 3.33 & 4.00 & 3.33 & 3.33 & 4.67 & 3.67 & 1.00 & 4.30 \\

English Pivot Thought & 4.00 & 3.67 & 13.00 & 2.67 & 2.33 & 3.67 & 6.67 & 4.67 & 3.33 & 3.67 & 4.80 \\

\noalign{\vskip 1mm}  
\hline
\hline
\noalign{\vskip 1mm}  
DIP w/o EP w/o CT & 3.33 & 5.00 & 4.67 & 3.67 & 2.00 & 4.00 & 1.67 & 3.67 & 1.33 & 4.00 & 2.73 \\

DIP w/ EP w/o CT & 7.33 & 4.00 & 8.00 & 4.33 & 3.67 & 2.67 & 3.33 & 4.33 & 2.00 & 4.67 & 4.20 \\

DIP & 7.00 & \textbf{3.67} & 12.00 & \textbf{3.67} & \textbf{5.33} & \textbf{4.33} & \textbf{7.00} & \textbf{5.00} & \textbf{4.33} & \textbf{7.00} & \textbf{5.83} \\
\noalign{\vskip 1mm}  
\hline
\end{tabular}
\caption{\label{tllama1}
Results for Llama-3.2 on SVAMP on 10 randomly selected low-resourced languages from FLORES-200. EP denotes the English pivoting translation process, and CT denotes chain-of-thought reasoning steps.
}
\end{table*}
\begin{table*}[thb!]
\scriptsize
\centering
\setlength\tabcolsep{3pt}
\setlength\extrarowheight{4pt}
\begin{tabular}{lcccccccccc|c}
\hline
\noalign{\vskip 1mm}  
\textbf{Model} & \textbf{kaz\_Cyrl} & \textbf{nso\_Latn} & \textbf{srp\_Cyrl} & \textbf{xho\_Latn}& \textbf{ibo\_Latn}& \textbf{tum\_Latn}& \textbf{asm\_Beng}& \textbf{bug\_Latn}& \textbf{ckb\_Arab}&     \textbf{azb\_Arab} & \textbf{Average}\\
\noalign{\vskip 1mm}  
\hline
\hline
\noalign{\vskip 1mm}  
Standard Prompting & 28.00 & 14.00 & 36.00 & 16.33 & 8.33 & 12.33 & 28.33 & 17.00 & 27.67 & 20.67 & 18.97 \\

Non-insertion Prompting & 25.00 & 23.00 & 30.67 & 27.00 & 23.33 & 22.67 & 21.00 & 29.00 & 25.00 & 24.00 & 22.67 \\

English Pivoting & 35.00 & 10.33 & 51.33 & 12.00 & 7.67 & 10.33 & 38.00 & 18.00 & 30.67 & 24.00 & 22.43 \\

English Pivot Thought & 22.67 & 5.67 & 27.33 & 10.33 & 6.33 & 9.00 & 14.00 & 18.00 & 20.33 & 15.33 & 12.97 \\

\noalign{\vskip 1mm}  
\hline
\hline
\noalign{\vskip 1mm}  
DIP w/o EP w/o CT & 22.67 & 17.00 & 32.67 & 29.00 & 21.33 & 25.33 & 18.00 & 39.67 & 18.00 & 24.33 & 22.80 \\

DIP w/ EP w/o CT & 27.33 & 17.67 & 31.00 & 36.00 & 18.33 & 29.67 & 17.00 & 38.33 & 31.33 & 40.67 & 28.00 \\

DIP & \textbf{44.00} & \textbf{29.00} & \textbf{50.33} & \textbf{48.00} & \textbf{32.00} & \textbf{39.67} & \textbf{39.00} & \textbf{55.67} & \textbf{48.33} & \textbf{51.67} & \textbf{43.27} \\

\noalign{\vskip 1mm}  
\hline
\end{tabular}
\caption{\label{tmixtral1}
Results for Mixtral on SVAMP on 10 randomly selected low-resourced languages from FLORES-200. EP denotes the English pivoting translation process, and CT denotes chain-of-thought reasoning steps.
}
\end{table*}
\begin{itemize}
\setlength\itemsep{0em}
\item \textbf{Standard Prompting} that directly asks the English-centric LLMs to answer the questions written in those non-English languages.
\item \textbf{Non-insertion Prompting} prepends/appends the dictionary to the prompt.
\item \textbf{English Pivoting} that asks the model to translate the question into English before answering \citep{kim-etal-2019-pivot}.
\item \textbf{English Pivot Thought} that asks the model to translate the question into English before answering with chain-of-thought reasoning.
\end{itemize}
\begin{table*}
\small
\centering
    \setlength\tabcolsep{3.7pt}
    \setlength\extrarowheight{4pt}
\begin{tabular}{l|cccc|ccc}
\hline
\noalign{\vskip 1mm}  
\textbf{Methods} & \textbf{\# improved} & \textbf{> 5 points} & \textbf{> 10 points} & \textbf{> 20 points} & \textbf{\# degraded} & \textbf{> 5 points} & \textbf{> 20 points}\\
\noalign{\vskip 1mm}  
\hline
\noalign{\vskip 1mm} 
Non-insertion Prompting & 87/200 & 9/87 & 1/87 & 0/87 & 113/200 & 9/113 & 0/113  \\  
English Pivoting & 181/200 & 80/181 & 8/181 & 0/181 & 19/200 & 0/19 & 0/19 \\  
English Pivot Thought & 193/200 & 8/193 & 14/193 & 151/193 & 7/200 & 0/7 & 0/7  \\  
\noalign{\vskip 1mm}  
\hline
\hline
\noalign{\vskip 1mm}  
DIP w/o EP w/o CT & 193/200 & 84/193 & 32/193 & 0/193 & 7/200 & 0/7 & 0/7 \\  
DIP w/ EP w/o CT & 198/200 & 86/198 & 82/198 & 7/198 & 2/200 & 0/2 & 0/2 \\  
DIP & \textbf{200/200} & \textbf{1/200} & \textbf{2/200} & \textbf{196/200} & \textbf{0/200} & \textbf{0/0} & \textbf{0/0} \\ 

\noalign{\vskip 1mm}  
\hline
\end{tabular}
\caption{\label{summm}
Statistics of the changes in accuracy with DIT and other baselines compared to Standard Prompting on GPT-4o with 200 languages on GSM8K. 100\% of the directions (200 out of 200) are improved with DIT. 
}
\end{table*}
\subsection{Datasets}
We construct synthetic benchmarks in 200 languages from FLORES-200 \citep{nllb2022} using the NLLB Translators from the following existing benchmarks:
\begin{itemize}
\setlength\itemsep{0em}
\item \textbf{GSM8K} is a benchmark of math word problems. We randomly sample 200 instances for each of the 200 languages from FLORES-200 for GSM8K. We also randomly sample 10 low-resourced languages from GSM8K and conduct experiments on the full 1,319 test instances on them \citep{2021arXiv211014168C}.
\item \textbf{SVAMP} is a benchmark of math word problems with varying structures. We randomly sample 10 low-resourced languages from FLORES-200 and conduct experiments on the full 1,000 test instances \citep{patel-etal-2021-nlp}.
\item \textbf{AQuA} is a dataset of algebraic word problems. We randomly sample 10 low-resourced languages from FLORES-200 and conduct experiments on their full.
\item \textbf{Date and Sport} We selected two specialized evaluation sets from the BIG-bench initiative \citep{2022arXiv220604615S}: Date Understanding, which requires inferring a date based on a given context, and Sports Understanding, which involves assessing whether a sports-related sentence is plausible or implausible. We randomly sample 10 low-resourced languages from FLORES-200 and conduct experiments on their full test set.
\end{itemize}
\begin{table*}[thb!]
\scriptsize
\centering
\setlength\tabcolsep{3pt}
\setlength\extrarowheight{4pt}
\begin{tabular}{lcccccccccc|c}
\hline
\noalign{\vskip 1mm}  
\textbf{Model} & \textbf{kaz\_Cyrl} & \textbf{nso\_Latn} & \textbf{srp\_Cyrl} & \textbf{xho\_Latn}& \textbf{ibo\_Latn}& \textbf{tum\_Latn}& \textbf{asm\_Beng}& \textbf{bug\_Latn}& \textbf{ckb\_Arab}&     \textbf{azb\_Arab} & \textbf{Average}\\
\noalign{\vskip 1mm}  
\hline
\hline
\noalign{\vskip 1mm}  
Standard Prompting&47.20&47.20&53.60&54.40&61.60&47.20&55.60&45.20&46.40&46.00&50.44\\
Non-insertion Prompting&48.00&39.20&44.40&43.60&43.60&42.40&43.20&46.00&41.60&43.60&43.56\\
English Pivoting&49.60&42.80&45.20&44.80&49.60&48.00&47.20&43.20&42.80&43.20&45.64\\
English Pivot Thought&\textbf{74.80}&58.00&63.60&64.00&73.20&52.40&69.60&40.40&56.00&63.60&61.56\\
\noalign{\vskip 1mm}  
\hline
\hline
\noalign{\vskip 1mm}  
DIP w/o EP w/o CT &48.00&38.00&44.40&45.60&47.60&40.40&41.60&38.40&39.20&44.80 & 42.80\\
DIP w/ EP w/o CT &49.60&44.80&46.80&48.00&50.40&48.80&45.60&45.20&44.00&46.80&47.00\\
DIP &72.40&\textbf{66.80}&\textbf{73.20}&\textbf{71.6}0&\textbf{76.00}&\textbf{66.40}&\textbf{77.60}&\textbf{70.00}&\textbf{75.60}&\textbf{75.60}&\textbf{72.52}\\
\noalign{\vskip 1mm}  
\hline
\end{tabular}
\caption{\label{t3}
Results for GPT-4o on Date Understanding on 10 randomly selected low-resourced languages. EP denotes the English pivoting translation process, and CT denotes chain-of-thought reasoning steps.
}
\end{table*}
\begin{table*}[thb!]
\scriptsize
\centering
\setlength\tabcolsep{3pt}
\setlength\extrarowheight{4pt}
\begin{tabular}{lcccccccccc|c}
\hline
\noalign{\vskip 1mm}  
\textbf{Model} & \textbf{kaz\_Cyrl} & \textbf{nso\_Latn} & \textbf{srp\_Cyrl} & \textbf{xho\_Latn}& \textbf{ibo\_Latn}& \textbf{tum\_Latn}& \textbf{asm\_Beng}& \textbf{bug\_Latn}& \textbf{ckb\_Arab}&     \textbf{azb\_Arab} & \textbf{Average}\\
\noalign{\vskip 1mm}  
\hline
\hline
\noalign{\vskip 1mm}  
Standard Prompting&49.60&56.00&42.80&56.80&55.60&53.20&43.60&54.00&27.20&45.20&48.40\\
Non-insertion Prompting&48.40&47.60&46.00&51.60&50.00&52.40&41.60&48.80&52.80&54.40&49.36\\
English Pivoting&48.40&48.80&50.80&48.00&42.80&47.60&52.40&47.60&51.60&45.60&48.36\\
English Pivot Thought&47.20&50.40&46.80&46.80&46.00&47.20&47.60&46.80&46.80&46.00&47.16\\
\noalign{\vskip 1mm}  
\hline
\hline
\noalign{\vskip 1mm}  
DIP w/o EP w/o CT &46.80&50.00&46.80&55.20&56.80&56.00&42.80&56.80&22.00&52.40&48.56\\
DIP w/ EP w/o CT &56.40&54.80&64.00&60.40&60.40&\textbf{60.40}&56.40&54.80&58.80&54.40&58.08\\
DIP &\textbf{58.80}&\textbf{64.40}&\textbf{67.20}&\textbf{60.80}&\textbf{62.40}&59.60&\textbf{58.80}&\textbf{64.00}&\textbf{63.20}&\textbf{59.20}&\textbf{61.84}\\
\noalign{\vskip 1mm}  
\hline
\end{tabular}
\caption{\label{t4}
Results for GPT-4o on Sports Understanding on 10 randomly selected low-resourced languages. EP denotes the English pivoting translation process, and CT denotes chain-of-thought reasoning steps.
}
\end{table*}
\subsection{Evaluation Metrics} 
Accuracy is used to evaluate the reasoning tasks. In addition, we use BLEU \citep{papineni-etal-2002-bleu} and chrF++ \citep{popovic-2015-chrf} to evaluate the translation quality as well as the intermediate thinking process. We use the evaluations provided by the sacreBLEU repository with the default signatures.\footnote{https://github.com/mjpost/sacrebleu}
\subsection{Synthetic Generation Quality}
Under our setting, almost all the words from the question prompt are assigned a dictionary. In order to ensure the quality of the generated dataset with the 10 languages we study, we employ three experienced human annotators. They are all postgraduate student who are studying for English-relevant degrees.  We use GPT to translate the synthetic benchmark back to English and ask them to annotate whether the translated-back English is the same as the original English. With this method, we only preserve the instances that are perfectly agreed upon by all the annotators, and the meanings are preserved. More than 90\% of the instances are preserved for each language, and this process ensures a perfect agreement for the final datasets among our annotators.
\section{Math Reasoning Tasks}
\subsection{GSM8K}

\paragraph{GSM8K} Table \ref{t1} presents the results on GSM8K written in 10 randomly selected low-resourced languages from FLORES-200 \citep{nllb2022} on GPT-4o. The results indicate the effectiveness of DIP in comparison to the baselines. English pivoting does help to improve the Standard Prompting baseline from an average of 11.04 to 17.37. Chain-of-thought reasoning is especially useful, improving the average score from 17.37 on English Pivoting to 44.88 on English Pivot Thought. Compared to the most naive Standard Prompting baseline, DIP obviously improves the average performance from 11.04 to 61.92. On bug\_Latn, DIP still obviously improves the strongest baseline English Pivot Thought from 20.85 to 60.50. We also observe the effectiveness of interleaving the dictionary by insertion, which improves traditional appending/prepending previously employed by machine translation, improving the average score from 10.02 on Non-insertion Prompting to 16.35 on DIP w/o EP w/o CT. We postulate such an interleaving manner makes it easy to catch the context between the mapped dictionary pairs, which enhances the model understanding.
\par
Due to space reasons, we leave Figure \ref{gsm8k_f} in the Appendix, which visually presents the performance gap between DIP and the baselines on the full 200 languages in FLORES-200 on GPT-4o. We observe that DIP has quite impressive improvements on lower-resource languages such as the ones in the fourth row. The improvements with DIP on higher-resourced languages, yet, there are still improvements. Table \ref{summm} presents the detailed improvement statistics, and we observe that while traditional English Pivoting and English Pivot Thought make good improvements, they are usually not drastic, with about a 5-10 points increase in accuracy. In comparison, DIP gives a large improvement with 196/200 languages enjoying an improvement with over 20 points in accuracy. This concludes the usefulness of DIP.
\subsection{SVAMP} Table \ref{t2} presents the results on SVAMP written in 10 randomly selected low-resourced languages from FLORES-200 \citep{nllb2022} on GPT-4o. Similar to the results on GSM8K, there is only one language in which DIP scored less than the English Pivot Thought baseline. On SVAMP, the chain-of-thought reasoning improves less (from 52.23 on English Pivoting to 58.70 on English Pivot Thought) than English pivoting (from 41.17 on Standard Prompting to 52.23 on English Pivoting). This means that the usefulness of these two techniques varies on different tasks. The overall improvement from DIP is obvious, where DIP improves the average score from 58.70 on English Pivot Thought to 73.50. We also observe the effectiveness of interleaving the dictionary by insertion, which improves Non-insertion Prompting, improving the average score from 41.47 on Non-insertion Prompting to 57.73 on DIP w/o EP w/o CT.
\par
\begin{table*}[thb!]
\scriptsize
\centering
\setlength\tabcolsep{3pt}
\setlength\extrarowheight{4pt}
\begin{tabular}{lcccccccccc|c}
\hline
\noalign{\vskip 1mm}  
\textbf{Model} & \textbf{kaz\_Cyrl} & \textbf{nso\_Latn} & \textbf{srp\_Cyrl} & \textbf{xho\_Latn}& \textbf{ibo\_Latn}& \textbf{tum\_Latn}& \textbf{asm\_Beng}& \textbf{bug\_Latn}& \textbf{ckb\_Arab}&     \textbf{azb\_Arab} & \textbf{Average}\\
\noalign{\vskip 1mm}  
\hline
\hline
\noalign{\vskip 1mm}  
Standard Prompting & 17.20 & 16.00 & 22.80 & 19.20 & 22.00 & 20.40 & 11.20 & 21.60 & 13.20 & 12.80 & 17.22 \\

Non-insertion Prompting & 16.00 & 14.40 & 19.20 & 16.80 & 13.20 & 12.00 & 15.60 & 16.40 & 13.60 & 15.20 & 15.16 \\

English Pivoting & 20.80 & 23.20 & 18.80 & 24.40 & 25.60 & 18.40 & \textbf{24.40} & 20.00 & 20.00 & 22.80 & 21.86 \\

English Pivot Thought & \textbf{21.20} & 20.40 & 17.20 & 24.40 & \textbf{25.60} & 18.80 & 18.80 & 18.80 & 22.40 & \textbf{24.40} & 22.16 \\

\noalign{\vskip 1mm}  
\hline
\hline
\noalign{\vskip 1mm}  
DIP w/o EP w/o CT & 14.40 & 20.40 & 16.80 & 17.20 & 20.40 & 16.80 & 10.40 & 17.20 & 18.00 & 18.40 & 19.16 \\

DIP w/ EP w/o CT & 20.40 & 23.60 & \textbf{22.80} & 19.20 & 20.80 & 16.00 & 18.40 & 15.20 & 21.20 & 19.60 & 21.70 \\

DIP & 20.40 & \textbf{26.80} & 20.00 & \textbf{29.20} & 24.00 & \textbf{22.00} & 19.20 & \textbf{22.80} & \textbf{23.60} & 20.40 & \textbf{23.94} \\
\noalign{\vskip 1mm}  
\hline
\end{tabular}
\caption{\label{t3llama}
Results for Llama-3.2 on Date Understanding on 10 randomly selected low-resourced languages. EP denotes the English pivoting translation process, and CT denotes chain-of-thought reasoning steps.
}
\end{table*}
\begin{table*}[thb!]
\scriptsize
\centering
\setlength\tabcolsep{3pt}
\setlength\extrarowheight{4pt}
\begin{tabular}{lcccccccccc|c}
\hline
\noalign{\vskip 1mm}  
\textbf{Model} & \textbf{kaz\_Cyrl} & \textbf{nso\_Latn} & \textbf{srp\_Cyrl} & \textbf{xho\_Latn}& \textbf{ibo\_Latn}& \textbf{tum\_Latn}& \textbf{asm\_Beng}& \textbf{bug\_Latn}& \textbf{ckb\_Arab}&     \textbf{azb\_Arab} & \textbf{Average}\\
\noalign{\vskip 1mm}  
\hline
\hline
\noalign{\vskip 1mm}  
Standard Prompting & 22.40 & 31.60 & 24.80 & 29.60 & 34.80 & 32.40 & 18.80 & 28.80 & 25.60 & 21.60 & 30.92 \\

Non-insertion Prompting & 14.40 & 17.20 & 18.00 & 19.20 & 20.80 & 26.00 & 13.20 & 20.80 & 11.20 & 14.40 & 18.52 \\

English Pivoting & 14.40 & 18.80 & 15.60 & 19.60 & 16.80 & 18.00 & 9.20 & 23.60 & 12.00 & 13.60 & 14.68 \\

English Pivot Thought & 21.60 & 24.00 & 29.60 & 25.20 & 25.60 & 26.00 & 18.80 & 29.60 & 24.00 & 21.60 & 25.01 \\

\noalign{\vskip 1mm}  
\hline
\hline
\noalign{\vskip 1mm}  
DIP w/o EP w/o CT & 48.00 & 41.60 & 43.60 & 47.60 & 48.40 & 43.20 & 44.00 & \textbf{51.60} & 39.20 & 41.60 & 40.64 \\

DIP w/ EP w/o CT & 39.60 & 34.40 & 36.80 & 37.60 & 42.40 & 41.60 & 34.00 & 40.00 & 32.40 & 36.00 & 35.78 \\

DIP & \textbf{48.80} & \textbf{41.60} & \textbf{51.20} & \textbf{52.00} & \textbf{51.20} & \textbf{48.40} & \textbf{45.60} & 49.60 & \textbf{45.20} & \textbf{57.20} & \textbf{49.08} \\

\noalign{\vskip 1mm}  
\hline
\end{tabular}
\caption{\label{t4mixtral}
Results for Mixtral on Date Understanding on 10 randomly selected low-resourced languages. EP denotes the English pivoting translation process, and CT denotes chain-of-thought reasoning steps.
}
\end{table*}
\begin{table*}[thb!]
\tiny
\centering
\setlength\tabcolsep{2.7pt}
\setlength\extrarowheight{4pt}
\begin{tabular}{lcccccccccc|c}
\hline
\noalign{\vskip 1mm}  
\textbf{Model} & \textbf{kaz\_Cyrl} & \textbf{nso\_Latn} & \textbf{srp\_Cyrl} & \textbf{xho\_Latn}& \textbf{ibo\_Latn}& \textbf{tum\_Latn}& \textbf{asm\_Beng}& \textbf{bug\_Latn}& \textbf{ckb\_Arab}&     \textbf{azb\_Arab} & \textbf{Average}\\
\noalign{\vskip 1mm}  
\hline
\hline
\noalign{\vskip 1mm}  
English Pivoting & 47.70 / 69.28 & 40.94 / 60.11 & 56.04 / 76.59 & 49.25 / 67.31 & 46.74 / 65.09 & 22.35 / 43.43 & 41.93 / 64.48 & 26.68 / 52.47& 34.14 / 53.31 & 33.31 / 58.05 & 39.91 / 61.01 \\
\noalign{\vskip 1mm}  
\hline
\hline
\noalign{\vskip 1mm}  

DIP & \textbf{57.43 / 75.81} & \textbf{54.11 / 71.80} & \textbf{72.93 / 84.47} & \textbf{69.63 / 81.91} & \textbf{61.61 / 75.82} & \textbf{46.86 / 67.54} & \textbf{58.59 / 76.32} & \textbf{71.59 / 84.97} & \textbf{62.24 / 78.75} & \textbf{55.09 / 72.34}  & \textbf{61.00 / 76.97} \\

\noalign{\vskip 1mm}  
\hline
\end{tabular}
\caption{\label{engtrans}
Evaluations on translation quality on 10 randomly selected low-resourced languages on GSM8K on GPT-4o, evaluated in BLEU / chrF++ scores.
}
\end{table*}
\begin{table*}[thb!]
\tiny
\centering
\setlength\tabcolsep{2.8pt}
\setlength\extrarowheight{4pt}
\begin{tabular}{lcccccccccc|c}
\hline
\noalign{\vskip 1mm}  
\textbf{Model} & \textbf{kaz\_Cyrl} & \textbf{nso\_Latn} & \textbf{srp\_Cyrl} & \textbf{xho\_Latn}& \textbf{ibo\_Latn}& \textbf{tum\_Latn}& \textbf{asm\_Beng}& \textbf{bug\_Latn}& \textbf{ckb\_Arab}&     \textbf{azb\_Arab} & \textbf{Average}\\
\noalign{\vskip 1mm}  
\hline
\hline
\noalign{\vskip 1mm}  
English Pivot Thought & 4.99 / 21.32 & 4.58 / 20.24 & 6.13 / 23.10 & \textbf{3.73 / 19.34} & 5.19 / 21.80 & 3.65 / 17.61 & \textbf{6.13 / 23.48} & 4.45 / 19.46& 4.68 / 20.40 & \textbf{5.52} / 21.52 & 4.91 / 20.83 \\

\noalign{\vskip 1mm}  
\hline
\hline
\noalign{\vskip 1mm}  
DIP & \textbf{5.91 / 23.16} & \textbf{4.84 / 20.85} & \textbf{7.96 / 25.38} & 2.79 / 16.47 & \textbf{5.89 / 22.80} & \textbf{5.16 / 20.94} & 5.97 / 23.47 & \textbf{6.94 / 24.26} & \textbf{5.54 / 22.40} & 5.30 / \textbf{22.87} & \textbf{5.63 / 22.26} \\

\noalign{\vskip 1mm}  
\hline
\end{tabular}
\caption{\label{thinking}
Evaluations on thinking quality on 10 randomly selected low-resourced languages on GSM8K on GPT-4o, evaluated in BLEU / chrF++ scores.
}
\end{table*}
Due to limited computational resources, we conduct experiments on SVAMP on Llama-3.2 in Table \ref{tllama1} and Mixtral in Table \ref{tmixtral1}, which shows that the results are consistent as shown on GPT-4o. Results on Llama-3.2 are usually lower than expected since those low-resourced languages are not perfectly supported on the Llama-3.2 we used.
\subsection{Ablation Study} The last three rows across all tables above indicate the effectiveness of different components in DIP, where the performance decreases when we remove English pivoting and chain-of-thought. We also observe that the performance of DIP variants is constantly better than or on par with Non-insertion Prompting. This concludes that the position of the dictionary is important, and with an interleaving dictionary only, DIP still surpasses some strong baselines, such as the English pivoting baseline as in Table \ref{t2} on SVAMP. 
\subsection{Commonsense Reasoning Tasks}
\paragraph{Date Understanding} Table \ref{t3} presents the results on Date Understanding written in 10 randomly selected low-resourced languages from FLORES-200 \citep{nllb2022} on GPT-4o. Similar to the results on math tasks, DIP is obviously better than all the baselines, except for on kaz\_Cyrl. We also note that the experiments on Llama-3.2 in Table \ref{t3llama} and Mixtral in Table \ref{t4mixtral} all give the same conclusion that DIP is obviously effective.
\par
On Llama-3.2, it is obviously better than the baselines, improving it from 17.22 on average on Standard Prompting to 23.94 on average with DIP. In contrast, on Mixtral, DIP gives a better improvement than on Llama, by to 49.08 on DIP, which probably indicate that DIP can gives a better improvement when the applied LLMs are stronger in performance. Nevertheless, the performance improvement from DIP is consistent. We also note that for some languages on Llama-3.2, DIP can be a bit lower than some baselines, such as in Table \ref{t3llama} on ibo\_Latn. However, this does not affect the overall final conclusion that DIP is obviously better than the baselines, according to the obvious averaged performance improvement with DIP as in Table \ref{t3llama}.
\paragraph{Sports Understanding} Table \ref{t4} presents the results on Sports Understanding written in 10 randomly selected low-resourced languages from FLORES-200 \citep{nllb2022}. Similar to the results on math tasks, DIP is obviously better than all the baselines, on all the languages. This suggests the consistent improvement and the usefulness of DIP in the task of Sports Understanding. 

\paragraph{Ablation Study}  The last three rows indicate the effectiveness of different components in DIP, where the performance decreases when we remove English pivoting and chain-of-thought. All those components are important to DIP.

\subsection{Translating Performance}
In order to study the effectiveness of DIP, we conduct deeper analysis as in Table \ref{engtrans} and Table \ref{thinking}. We found that there are two main reasons why DIP is useful. We found that DIP usually gives a better English pivoting performance as in Table \ref{engtrans}. This succeedingly gives a better thinking process under the evaluations as demonstrated in Table \ref{thinking}.
\par
As in Table \ref{engtrans}, we see that the translation performance has improved from 39.91 in BLEU to 61.00 in BLEU, which is a large improvement. This is also consistent with chrF++ scores, improving from 61.01 to 76.97. This represents the usefulness of DIP in terms of the intermediate translation.
\par
In Table \ref{thinking}, we see that while the thinking process can be diverse, there is a clear improvement between DIP and the baselines in terms of their thinking process to the original ground truth in GSM8K. This also indicates that the synthetic benchmark has reasonable quality, as the translated-back English aligns well with the original English dataset.
\section{Related Work}
\paragraph{Multilingual Tasks on Large Language Models} There has been limited research conducted on effective methods for prompting English-centric large language models on non-English tasks, such as the standard cross-lingual tasks such as machine translation. Most of the existing research focused on evaluating the translation performance of English-centric LLMs, using prompts such as `Translate to {language\_name}: {text}' \citep{NEURIPS2020_1457c0d6,2021arXiv211210668L,2022arXiv221105100W,2022arXiv220501068Z}. Different prompt formats are explored \citep{10.1145/3411763.3451760,2023arXiv230304048W}, Furthermore, \citet{2022arXiv220211822G} have investigated the potential need for prompts for regulating the formality or specific dialect of the generation. Finally, \citet{2022arXiv221202437A} and \citet{2022arXiv221109102V} focused on identifying appropriate in-context examples to improve machine translation quality with LLMs. In addition to machine translation which has already scaled to over 200 languages from FLORES-200 \citep{nllb2022}, there is also a trend in solving non-English reasoning tasks on English-centric LLMs \citep{huang-etal-2023-languages}, yet, the number of languages studied are usually insufficient, with about tens of languages.
\paragraph{Dictionary-based Method for Multilingual Language Models} 
This research is relevant to the idea of lexical restrictions in the task of machine translation. This can be divided into either hard constraints \citep{hokamp-liu-2017-lexically,post-vilar-2018-fast} or soft constraints \citep{2019arXiv190409107S,dinu-etal-2019-training,10.5555/3491440.3491936}.
\par
There have been several works that explored using dictionaries in supervised machine translation. \citet{2016arXiv161007272Z} enhance neural machine translation (NMT) by integrating a bilingual dictionary that incorporates less common or unseen words found in the bilingual training data. \citet{arthur-etal-2016-incorporating} improve the translation of rare words by augmenting the system with discrete translation lexicons and leveraging the attention vector to identify the relevant lexical probabilities. \citet{10.1145/3377713.3377801} employs a dictionary to generate synthetic parallel data, thereby enhancing the training of NMT models. While most of previous work has focused on using dictionaries for the task of machine translation, doing multilingual reasoning tasks is under-studied.
\par
In contrast, DIP is the first work that exploits the use of a dictionary in terms of reasoning tasks in non-English languages on English-centric LLMs.
\section{Conclusions}
In conclusion, our proposed method, DIP, provides an effective solution to multilingual reasoning by inserting English counterparts for non-English prompts, experimental analysis indicates that DIP enhances translation accuracy and reasoning capabilities within English-centric LLMs. Through extensive experiments on approximately 200 languages using synthetic multilingual benchmarks created from existing benchmarks such as GSM8K and SVAMP, DIP has demonstrated substantial improvement in multilingual math and commonsense reasoning tasks across various LLMs. We also found that interleaving the dictionaries plays an important factor in the final performance.
\section*{Limitations}
This paper presents an analysis of 200 languages only. However, there are more than thousands of languages around the world. The paper can be further extended by including more languages as well as more analysis.
\section*{Ethical Statement}
We honour and support the ACL ARR Code of Ethics. There is no ethical issue known to us. Well-known and widely used LLMs are used in our work, which is subjected to generating offensive context. However, the above-mentioned issues are widely known to commonly exist for LLMs. Any content generated does not reflect the view of the authors.
\bibliography{custom}
\appendix
 \begin{figure*}[t!]
\begin{center}
\vspace{0mm}
\centerline{
\includegraphics[width=15.5cm]{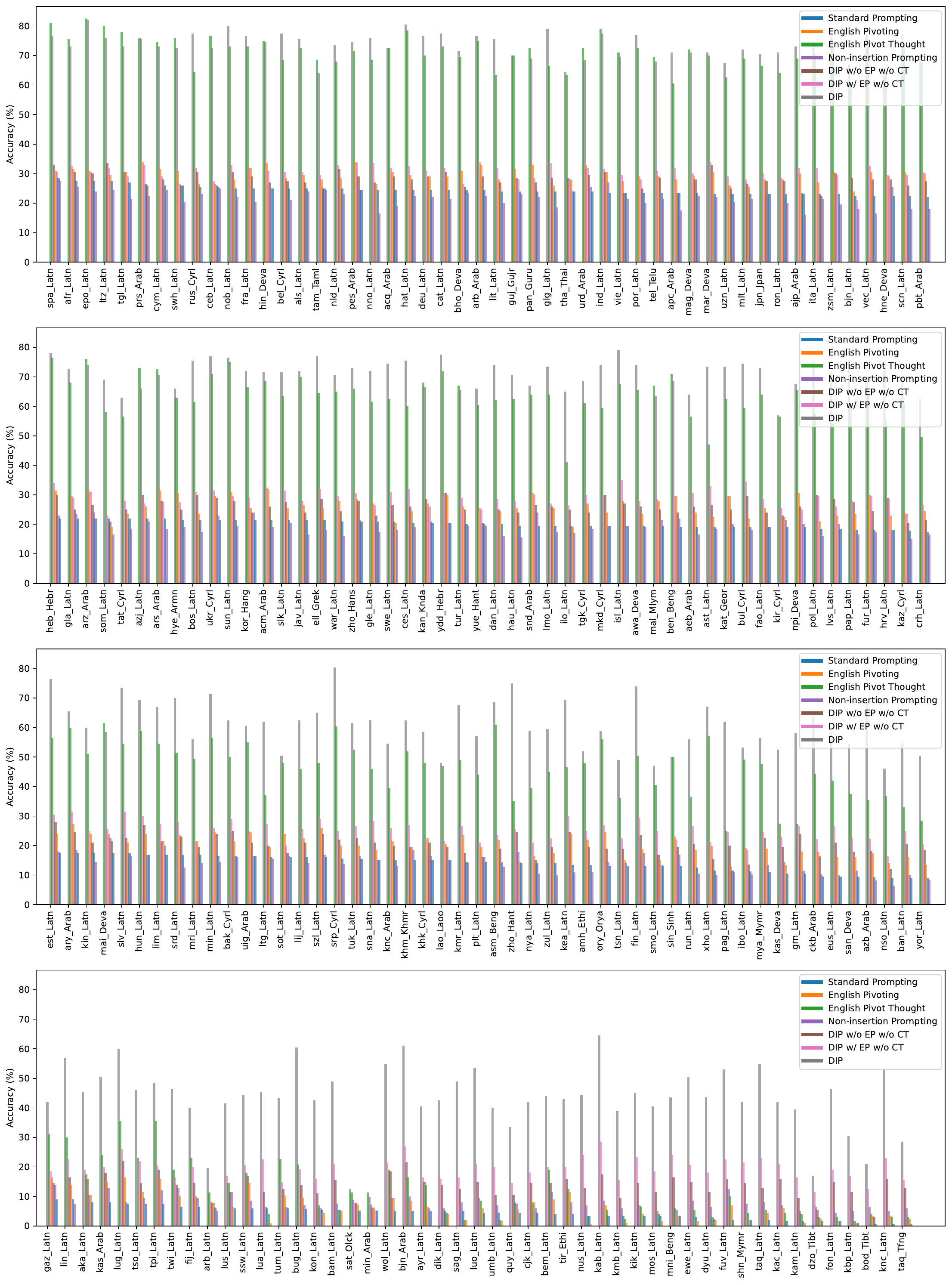}}
\caption{An illustrated comparison of the GSM8K dataset with 200 languages from FLORES-200 on six baselines/variants and DIT. While DIP has good improvements in the higher-resourced languages, the performance improvement of DIT is especially obvious in low-resource languages in the last row of the graph. EP denotes the English pivoting translation process, and CT denotes chain-of-thought reasoning steps.}
\label{gsm8k_f}
\end{center}
\vspace{-5mm}
\end{figure*}

\end{document}